# The Syntax of *qulk*-clauses in Yemeni Ibbi Arabic: A Minimalist Approach

Zubaida Mohammed Albadani
Department of English, Qalam University, Yemen
Mohammed Q. Shormani
Department of English & Translation, Ibb University, Yemen
shormani@ibbuniv.edu.ye/https://orcid.org/0000-0002-0138-4793

December, 2025

**Abstract**
This study investigates the syntax of *qulk*-clauses in Yemeni Ibbi Arabic (YIA) within the Minimalist Program. The construction *qulk*, a morphologically fused form meaning 'I said,' introduces embedded declarative, interrogative, and imperative clauses, often without an overt complementizer. The central proposal of this study is that *qulk*-clauses are biclausal structures in which *qulk* functions as a clause-embedding predicate selecting a full CP complement. By applying core minimalist operations, viz., *Merge, Move, Agree*, and *Spell-out*, the study provides a layered syntactic analysis of *qulk*-clauses, illustrating how their derivation proceeds through standard computational steps and post-syntactic processes such as Morphological Merger. The proposal also accounts for dialect-specific features like bipartite negation, cliticization, and CP embedding. The findings offer theoretical contributions to generative syntax, specifically minimalism. The study concludes raising theoretical questions concerning extending the analysis to the addressee-clause *qil-k* 'you said'. It also provides insights into the possibility of the universality of minimalism.

**Keywords:** Yemeni Ibbi Arabic, *qulk*-clauses, Minimalist Program, biclausal structures, bipartite negation

## 1. Introduction

The syntax of embedded clauses in spoken Arabic dialects has received increasing attention in generative linguistics (Aoun et al., 2010), particularly within the Minimalist Program (MP) framework (Chomsky, 1995 et seq). Despite this growing interest, specific constructions in less-studied dialects such as *qulk*-clauses in Yemeni Ibbi Arabic (YIA), remain underexplored. Clause embedding is a core phenomenon in syntactic theory, reflecting how human languages structure complex propositions and manage information hierarchies. Within the MP, embedded clauses are typically introduced by complementizers and selected by matrix predicates in a structurally layered architecture. While the syntax of embedded clauses has been extensively studied in Standard Arabic and major world languages, much less is known about how such structures manifest in lesser-documented Arabic dialects, particularly those spoken in Yemen, as Yemen Arabic, in general, is an underinvestigated language variety (Shorman, 2019).

One such construction of particular interest is the *qulk*-clause in YIA, a variety of Yemeni Arabic spoken in the Ibb region. The form *qulk* (literally, I said) is a morphologically fused unit combining the verb *qul* 'say' and a subject clitic *-k* 'I'. This construction introduces embedded declarative, interrogative, and imperative clauses that often lack overt complementizers but still exhibit full sentential structure. These *qulk*-clauses are not only frequent in daily speech but are also syntactically rich, involving biclausal configurations, cliticization, and dialect-specific patterns of negation and word order.

Thus, this study aims to provide a minimalist syntactic analysis of *qulk*-clauses in YIA within the PM, and see whether the minimalist assumptions account for such a dialect. The core proposal is that *qulk*-clauses are biclausal in nature, in which *qulk* functions not merely as a lexical verb but as a clause-embedding predicate that selects a full CP (=Complementizer Phrase) as its syntactic complement. This proposal is motivated by evidence from clause structure, agreement patterns, negation, and word order, all of which indicate that the embedded clauses behave as independent sentential domains subject to universal grammatical constraints. In developing this proposal, the study applies fundamental minimalist operations, viz., *Merge, Move, Agree*, and *Spell-out*, to demonstrate how *qulk*-clauses are derived within the internal computational system of the human brain. Additionally, the analysis incorporates post-syntactic processes such as Morphological Merger, which accounts for the surface realization of clitics and verb forms. These operations are shown to apply in consistent and predictable ways across the range of *qulk*-clause types, whether declarative, interrogative, or imperative.

Beyond contributing to syntactic theory, the study has broader implications for Arabic dialectology and linguistic typology. It provides empirical evidence that dialectal Arabic varieties like YIA exhibit rich structural complexity, enhancing the dialectical evidence of minimalism. challenging assumptions that spoken varieties are syntactically impoverished. Moreover, the analysis illustrates how language-specific parametric variation, such as bipartite negation (*ma…-š*), null subject licensing, and word-order flexibility, interacts with universal principles posited by the Minimalist framework. This study, thus, advances our understanding of how *qulk*-clauses are represented, processed and interpreted in the human brain (of YIA speakers) and offers a theoretically grounded model that can inform future work on other understudied dialects. Given existing studies on the syntax of YIA conducted by (Shormani, 2017, 2021; Shormani & Alhussen, 2024; Shormani & Qarabesh, 2018), this study addresses a gap by tackling the syntax of *qulk*-clauses in this spoken dialect, and adds valuable insights into how minimalism can be extended to account for syntactic phenomena of such dialects. In other words, through original data and minimalist analysis, this study offers new insights into how spoken dialects instantiate universal syntactic principles, contributing to ongoing discussions on the interface issues correlating the syntax and discourse.

Thus, this article is organized as follows. Section 2 spells out the theoretical foundations of the study, outlining the development of generative grammar from Standard Theory to the MP, and establishing the relevance of minimalist assumptions to Arabic syntax. It also reviews the relevant literature on clause structures in Arabic and other languages, identifying gaps and setting the stage for the current analysis. Section 3 introduces the central proposal of the paper, presenting a detailed derivation of *qulk*-clauses within the Minimalist framework and arguing for their biclausal structure with CP-embedding. Section 4 presents and analyzes empirical data from YIA, covering declarative, interrogative, and imperative *qulk*-clauses and illustrating how minimalist operations account for their syntactic behavior. Section 5 discusses the theoretical and typological implications of the analysis, particularly in relation to Universal Grammar (UG) and dialectal variation. Section 6 concludes the paper, providing some recommendations for future research.

## 2. Theoretical Foundations

We argue that studying *qulk*-clauses in Yemeni Ibbi Arabic is best handled within the PM, the latest generative framework, proposed by Chomsky (1995, et seq). This intellectual trajectory originates with early grammatical studies by ancient scholars like Pāṇini and Aristotle, extends through the systematic analyses of medieval grammarians such as Priscian and Donatus (Robins, 2013), and culminates in contemporary biolinguistic approaches. The cognitive revolution of the 1950s marked a pivotal transition toward mentalistic explanations of language, fundamentally shaped by Chomsky's (1957) *Syntactic Strictures* introducing generative grammar and the concept of an innate Universal Grammar (UG) (Ouhalla, 1999; Shormani, 2013).

Chomsky's Standard Theory (ST) (1957) introduced kernel sentences—simple, active declaratives generated via Phrase Structure Rules (PSRs) such as in (1):

(1) S → NP VP

The internal lexical items in () are then undergone transformational processes, known as transformational Rules (TRs), such as passivization (Chomsky, 1957; Ghulfan, 2010; Shormani, 2017). Criticisms of PSRs included their structurally "unfair" representations, like flat hierarchies conflating complements and modifiers, and their generation of semantically anomalous sentences as in (2), from Shormani (2024a).

(2) *Ali gave Alia a mountain.

Structures like (2) necessitate developing the then syntactic Theory, whose results came up with the Modified Standard Theory (MST) replacing kernels with Deep Structure (DS), where PSRs operate, and Surface Structure (SS), derived via TRs, integrating semantics (Shormani, 2024a). The Extended Standard Theory (EST) introduced X-bar theory (Jackendoff, 1977), leading to the Principles and Parameters (P&P) framework (1980s). The P&P framework posits UG as innate principles such as obligatory subjects and language-specific parameters such as null subjects in Arabic (cf. Chomsky, 1995). It links linguistics to biology, arguing that UG is biologically hardwired, enabling rapid acquisition (Pinker, 1994; Shormani, 2013). The Minimalist Program (Chomsky, 1995, et seq) streamlines P&P, reducing syntax to efficient operations mediating sound—Phonetic Form (PF)—and meaning—Logical Form (LF). The Computational System of Human

Language ($C_{HL}$) employs the operation *Select* for lexical retrieval, *Merge* for hierarchical combination, *Move* for repositioning, *Agree* for feature matching, and *Spell-Out* for PF and LF interfaces (Chomsky, 2000; Carnie, 2021). MP's Strong Minimalist Thesis (SMT) posits language as an optimal interface system, minimizing computational complexity (Berwick & Chomsky, 2016; Zwart, 1998).

## 2.1. Literature Review

The study of clause structure cross-linguistically within the framework of generative linguistics has provided significant insights into syntactic phenomena across languages, including understudied dialects such as Yemeni Ibbi Arabic. Shormani (2024b) highlights the profound influence of Generative Linguistics (GL) on contemporary linguistic theory, particularly in its application to computational models of language via artificial intelligence (AI). While AI systems such as Large Language Models (LLMs), have adopted GL's rule-based approaches to syntax, fundamental differences remain, particularly in their reliance on textual data as opposed to the multimodal input inherent to human language acquisition. This distinction underscores the necessity of empirical investigations into natural language structures, including dialect-specific constructions like *qulk*-clauses in YIA.

The implementation of minimalist syntactic principles in clause analysis has generated significant theoretical advancements. Al-Samki's (2018) investigation of relative clauses in Arabic and English, conducted through the lens of Phase Theory, provides a substantive critique of conventional promotion analyses while simultaneously furnishing analytical methodologies pertinent to the examination of embedded constructions in YIA. These scholarly contributions collectively demonstrate the explanatory power of minimalist syntax in addressing cross-linguistic parametric variation, a crucial consideration in the investigation of the distinctive syntactic characteristics exhibited by *qulk*-clauses. Additionally, Al-Samki and Al-Ghrafy (2023) present a reconceptualization of English imperative constructions within the Minimalist Program framework, introducing the notion of an Imperative Phrase (ImpP) as a distinct functional projection. This theoretical model may prove instrumental in examining imperative structures incorporated within *qulk*-clauses in YIA. In a related vein.

The syntactic analysis of Arabic dialects has increasingly incorporated minimalist principles. Alharbi (2017) simplifies the classification of Arabic copular clauses into two primary types—descriptive and equality sentences— demonstrating how definiteness and case assignment interact with syntactic movement. These findings bear relevance to the structure of *qulk*-clauses, where matrix verbs such as *qulk* govern the syntactic behavior of embedded clauses. Guella’s (2010) investigation of Maghrebi Arabic relative clauses further illustrates dialectal innovation, noting the replacement of Classical Arabic’s gendered relative pronouns with the neutral elli. This streamlining mirrors YIA’s use of the fixed negation marker *mā...š* in *qulk*-constructions, suggesting a broader trend toward morphological simplification in spoken Arabic varieties.

Comparative syntactic research offers crucial theoretical foundations for analyzing clause alternation patterns. Rohdenburg's (2019) corpus-based study refutes the purported diachronic dominance of gerundive constructions over finite that-clauses in English, revealing instead their sustained prevalence in structurally complex contexts - a retention phenomenon mirroring YIA's preservation of full CP-embedding in *qulk*-clauses notwithstanding the dialect's general syntactic reduction tendencies. Complementing this structural perspective, Pryor (2007) advances a semantic distinction, arguing that that-clauses primarily function as specificational rather than referential elements - an analytical framework that similarly characterizes *qulk*-clauses as content-specifying rather than purely complementizing structures in YIA's syntactic architecture.

Despite these advances, the syntax of YIA remains underexplored, particularly with respect to embedded constructions. Al-Shami and Dkhissi (2021) examine tense and subject feature interactions in Yemeni Arabic more broadly, yet *qulk*-clauses specifically lack a dedicated minimalist treatment. Fakih’s (2003) comparative studies of Arabic dialects underscore the necessity of granular analyses, asserting that "each Arabic dialect presents unique syntactic features" (p. 34). This observation underscores the importance of the present study, which seeks to systematically analyze *qulk*-clauses within YIA, filling a critical gap in the literature.

Theoretical debates regarding clause function further contextualize this study. Storms (1966) associates that-clauses with the tripartite functions of language—communicative, expressive, and purposive—while Kac (1972) critiques performative analyses of causal clauses. These foundational works inform our understanding of *qulk*’s dual role as both a matrix verb and a discourse marker, bridging syntactic form and discursive function.

This synthesis reveals both the richness of existing syntactic theory and persistent gaps in Arabic dialectology. While Minimalism provides robust tools for analyzing clause structure and embedding (Chomsky, 1995; Shormani, 2024a), its application to understudied varieties like YIA remains limited. The current study addresses this gap by providing the first Minimalist analysis of *qulk*-clauses across declarative, interrogative, and imperative contexts, contributing to both linguistic theory and Arabic dialectology. By examining how YIA's structures align with or challenge universal principles, this study not only documents a unique grammatical phenomenon but also tests the boundaries of Minimalist syntax in accounting for linguistic diversity.

Tackling the syntax-discourse interface in the clausal structure, Shormani (2017) examines how word order variation, particularly SVO constructions, and the presence of silent or covert topics in Arabic affect the interpretation of referential *pro*. It argues that null pronouns pros in subject positions are not resolved purely through syntax *per se*, but also via the discourse components, such as topic continuity and prominence. The study shows that in SVO sentences, an overt subject often marks a topic shift, while in VS or pro-drop contexts, topic continuity is preserved. In this sense, the interpretation of *pro* is regulated at the syntax-discourse interface, where the TP structure interacts with information structure, viz., CP, discourse coherence, and referential accessibility. The study contributes to understanding how discourse-driven factors shape syntactic choices and pronoun interpretation in Arabic. Another study in which Shormani (2021) investigates another clausal structure is conducted to investigate the imperative structure in Yemeni Arabic. It tackles the structure and interpretation of imperatives in Arabic, emphasizing how syntactic derivation interacts with discourse functions at the syntax-discourse interface. It argues that Arabic imperatives are syntactically formed through verb movement and a null subject *pro*, (unlike PRO in English which lacks tense, and occurs in nonfinite clauses) projecting functional layers related to topicality and force. This study highlights that the imperative construction is not purely syntactic but deeply shaped by discourse roles, such as speaker authority, deontic modality, and speaker-hearer interactions as interlocuters. The syntax-discourse interface is shown to govern constraints on subject realization, and discourse appropriateness, revealing that imperatives serve as a testing ground for how grammatical form encodes discourse-related meaning in Arabic.

A further study examining interface properties has been conducted by Shormani and Qarabesh (2018) presenting an integrated analysis of vocative expressions in Arabic, arguing that they are not merely peripheral or pragmatic devices but occupy a legitimate place within the syntactic structure of clauses. Drawing on data primarily from Yemeni Arabic, the authors propose that vocatives are structurally situated in the left periphery of the clause, within a distinct Vocative Phrase (VocP) projection. This projection is part of the C-domain and functions in coordination with discourse mechanisms, particularly through establishing a link with an abstract or covert subject (pro) in imperatives. The paper shows that vocatives perform an aboutness-topic function, anchoring the clause to a specific discourse participant and thus fulfilling both syntactic and discourse roles. This dual function underscores the relevance of the syntax–discourse interface in analyzing vocatives, where formal syntactic operations and pragmatic interpretation work together to assign vocatives their communicative value. Ultimately, the study highlights the grammatical visibility of vocatives and challenges the traditional view that treats them as structurally marginal or entirely pragmatic. Additionally, Shormani and Qarabesh (2018) explore the grammatical status of vocatives by situating them at the intersection of syntax and discourse. They argue that vocatives are not merely pragmatic or peripheral elements but are syntactically encoded in the left periphery of the clause, occupying a dedicated projection (VocP) within the CP domain. Crucially, the study emphasizes that the interpretation of vocatives arises at the syntax–discourse interface, where formal syntactic structure interacts with discourse roles. Vocatives function as discourse anchors, identifying the addressee and establishing an aboutness-topic relation that links the clause to a specific participant in the discourse. This interface relation is made syntactically visible through coreference between the vocative and an implicit subject (pro) in imperative constructions. The paper thus demonstrates that vocatives contribute to both the formal derivation of clause structure and its communicative interpretation, supporting a view of grammar in which discourse functions are structurally represented and mediated through syntactic mechanisms.

Additionally, Shormani (2021) provides a detailed analysis of Arabic imperative clauses by situating them at the syntax–discourse interface, arguing that a complete understanding of these structures requires integrating their internal syntactic properties with discourse-level functions. He proposes that the interpretation of Arabic imperatives emerges from how syntax and discourse interact, particularly through performative uses that link propositional content with informational cues in discourse. Central to his account is the claim that the thematic subject of imperatives is a second-person null pronoun, namely *pro*, while any overt nominal appearing before the verb is not a subject in the traditional sense but functions as a topic in the C-domain that co-refers with the null *pro*(s). This co-referentiality is formalized as a kind of Agree with feature matching, giving rise to (non)local A'-chains. For core imperatives lacking an explicit topic, he postulates a null topic in Spec,TopP whose interpretation depends on discourse context, thereby showing how imperative syntax is shaped by, and interpreted through, discourse structures.

Finally, investigating negative imperatives in Yemeni Arabic, Shormani and Alhussen (2024) argue that these structures exhibit a bipartite negation structure combining a preverbal particle *lā* with a postverbal suffix *–š*, a pattern reminiscent of French negative imperatives. Syntactically, this arises through a split in the Negation head: NegP hosts *lā,* while NegClP hosts the clitic *–š*, reflecting their distinct syntactic positions and functions. The study integrates a phase-based syntactic framework with discourse-driven interface operations. Negative imperatives embed a covert 2nd-person subject (pro) in Spec,*v*P, and any overt prenominal element is not a subject but rather a topic occupying specifier of TopP. This topic is coreferential with the implicit subject, and their coreference is established via an Agree-as-Match relation at the syntax–discourse interface, ensuring full interpretability of *pro*. Through this analysis, Shormani and Alhussen demonstrate that bipartite negation in YA is not merely morphological; it is a consequence of syntactic structure shaped by discourse requirements. The partitioned negation head reflects both universal functional architecture and the necessity to interpret negation and topicalization as discourse salient features, thus providing a compelling case of how syntax and discourse mechanisms interact to yield the observed pattern.

Furthermore, Shormani and Alhussen's study ties this bipartite structure to the syntax–discourse interface, proposing that *mā...-š* contributes not only to grammatical well-formedness but also to discourse-level interpretation, especially in contexts requiring emphatic, contrastive, or categorical negation. For instance, the presence of both markers can signal the speaker's strong rejection of a proposition or reinforce the factuality of the negated event. Moreover, the positioning of *-š* after the verb aligns with the VSO word order dominating Arabic dialects, supporting the argument that even spontaneous speech is shaped by deep structural principles linked to UG. This approach shows how negation in YA exemplifies the interface between syntax derivation, employing the mental operations including *Select, Merge, Agree,* and *Spell-out* and discourse, mainly concerning interpretation via topic, and focus Bipartite negation, thus, serves as a key example of how a dialect-specific pattern, viz., *lā/mā–š* is constrained by and reveal underlying universal mechanisms at the syntax–discourse interface.

## 3. Proposal

Given our argument so far, this study posits that *qulk*-clauses in YIA should be handled within the PM, which gives such constructions enough room to account for their syntactic peculiarities. The central proposal is that *qulk*-clauses are biclausal in nature, in which the verb *qulk* 'I said' functions not merely as a lexical verb but as a clause-embedding predicate that selects a CP complement. qulk-clauses syntactic behavior in YIA demonstrates properties of functional heads introduce embedded propositions. The core structure of a *qulk*-clause can be represented in (3).

3 a. [$_{TP}$ *qulk* … [$_{TopP}$ Topic [$_{vP}$ ….. [$_{NegP}$ [$_{Neg}$ lā [$_{NegClP}$ [$_{NegCl}$ –š …]]]]]]] ➔ declarative/imperative

b. [$_{TP}$ *qulk* …. [$_{ForceP}$ wh-operator [$_{vP}$ ….. [$_{NegP}$ [$_{Neg}$ lā [$_{NegClP}$ [$_{NegCl}$ –š …]]]]]]] ➔ interrogative

For declarative and imperative *qulk*-clauses, we propose (3a) and for interrogative *qulk*-clauses, (3b) is proposed. In declarative and imperative *qulk*-clauses, the matrix clause is initiated with *qulk*-clause, introducing an embedded clause that begins with TopP. Cartographically, we assume that the matrix clause, which hosts *qulk*-clause is declarative in nature, specifically in our story. And that what encodes the clause type, viz., [+declarative], [+interrogative], or [+imperative] features, is the vP of the embedded clause in YIA, and is therefore taken as the highest functional projection of the embedded domain (Rizzi, 1997). TopP hosts an overt topic, TP carries tense and the overt subject (Rizzi, 1997; Chomsky, 2001). NegP and NegClP handle the bipartite negation (lā...š) (Shormani and Alhussen, 2024), while ForceP carries the [+interrogative} feature.

Based on the cartography of YIA in (3), and lexical insertion. *qulk* merges in V, raises to v and lands in T. Given these two processes, Agree comes to play. We define Agree as a mechanism for feature valuation. Agree established between T and the DP located in Spec-vP values the φ-features (e.g., person, number, gender) of T, hence interpreted and deleted. In the embedded clause, and given (3) either TopP or ForceP would be projected. As for the former, there two projections, viz., affirmative and negative. And given the bipartite negation in YIA, two other functional projections are necessitated, i.e. NegP and NegClP projected within the TP domain (Shormani & Alhussen, 2024). Adopting the theoretical assumptions of the MP (Chomsky 1995, 2000, 2001), this analysis seeks to explain the structural and derivational properties of these constructions by decomposing them into core syntactic layers, exploring the interaction between syntax and morphology, and accounting for the surface realization of elements.

Having these minimalist foundations, notions and conceptions, we now proceed with discussing our proposal, which will be detailed in the section to follow:

## 4. *qulk*-clauses in Yemeni Ibbi Arabic

Recall that *qulk*-clauses in YIA represent a distinctive syntactic phenomenon where the particle *qulk* (I said)—incorporating the bound subject clitic -k (1sg)—functions as both a matrix verb and a complementizer-like predicate introducing embedded clauses for reported speech. Unlike standard verbal constructions, this cliticized form structurally embeds a layered embedded clause that projects a reduced left periphery, composed of ForceP, TopP and FocP, followed by TP. This modified structure allows for clause typing and focus licensing while maintaining cross-linguistic consistency with universal complementation principles.

### 4.1. Declarative *qulk*-clauses

Declarative sentences express factual content and appear in three forms:

**Affirmative:**

(4) a. qul-k ʕali jāʔ
say-1.SG Ali come.PST.3.MS
'I said that Ali came'

b. qul-k ʔana ʔištaruk al-kitāb al-jadiid
say-1.SG I buy.PST.1.SG the-book the-new
'I said that I bought the new book'

c. qul-k lak ʔaḥmad ʔištari al-kitāb
say-1.SG to.you.MS Ahmad bought.PST.2.MS the-book
'I said to you: Ahmad, bought the book'

The facts in examples (4a–c) demonstrate that the verb *qul* 'say' introduces affirmative declarative clausal complements in YIA without an overt complementizer. In (4a), the embedded clause has a 3rd person subject *ʕali*, while (4b) shows a 1st person subject *ʔana*, indicating that the complement can shift reference depending on the reported speech. Crucially, all three cases lack an overt complementizer, implying that YIA allows for a structurally reduced CP, where only ForceP and TopP are activated, without projecting FocP, and without an overt complementizer—especially in affirmative declarative contexts. This interpretation aligns with the fundamental distinction in Rizzi's (1997) model where TopP is used in declarative sentences while ForceP is reserved for interrogative elements. This proposal aligns with Rizzi's cartographic model of the left periphery, while reflecting the cross-linguistic pattern in which verbs of saying license embedded clauses with null complementizers and a minimal CP structure (Hankamer & Mikkelsen, 2012; Haegeman, 2006).

Let us now consider the example in (5) and analyze how it can be derived using our proposal in (3a). (5) will have the derivation in (6):

(5) qul-k ʕali jāʔ
say-1.SG Ali come.PST.3.MS
'I said that Ali came.'

(6)

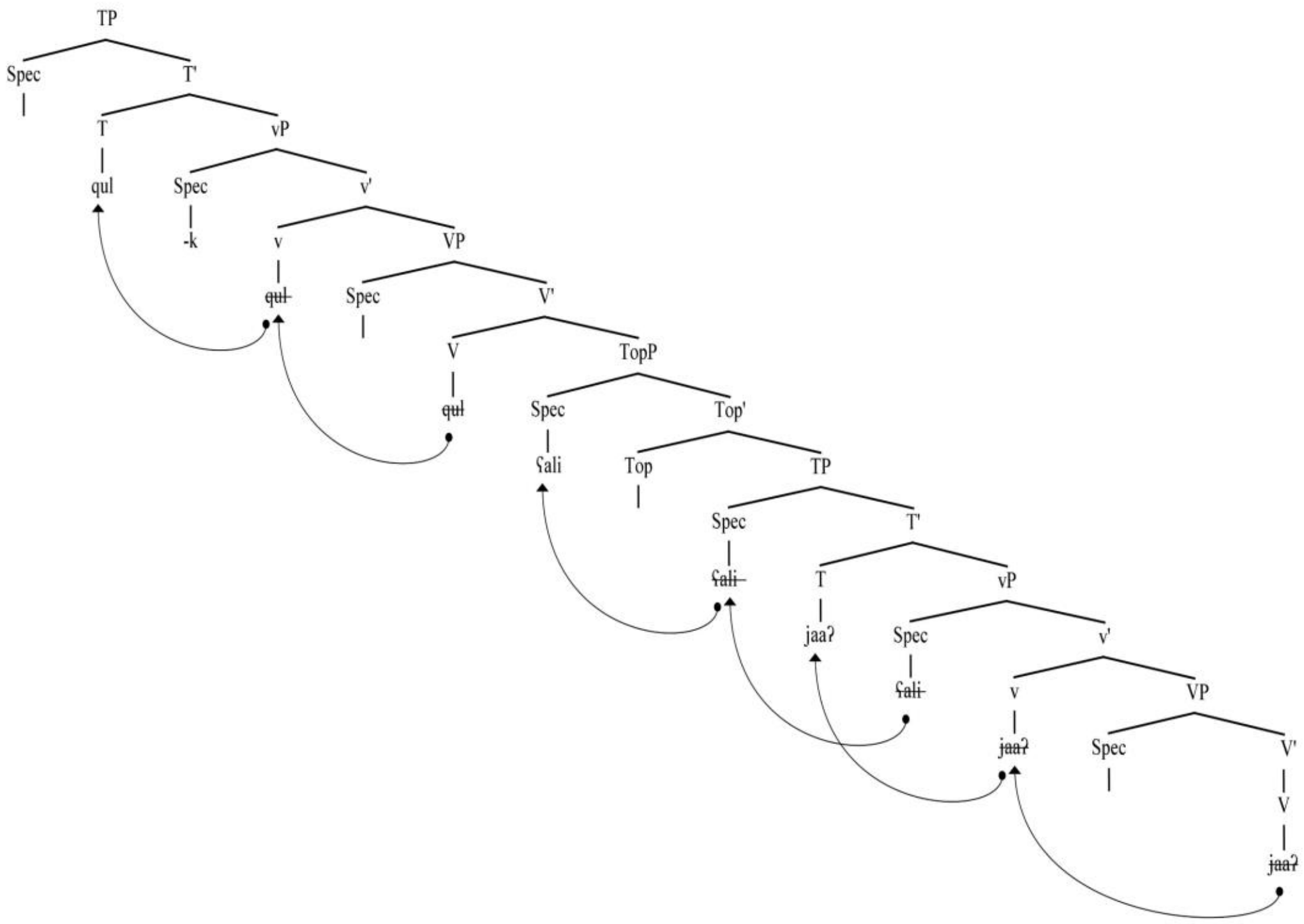

Details aside, the matrix clause forms first: the verb *qul* originates in its base position V and then moves to the tense position T to acquire past tense morphology and agreement features. The subject pronoun "-k" (representing "I") starts as an independent element in its own position (Spec-vP) but cannot remain separate—it needs to attach to the verb. As the verb *qul* moves to the tense position T, the subject pronoun *-k* jumps and clings to it, forming the fused word *qul-k*. Once the matrix domain is completed, the embedded declarative clause is constructed. The subject *ʕali* merges with the lexical verb jāʔ "came", which raises to T to satisfy tense features. In accordance with this proposal, since the clause is declarative and lacks clause-typing requirements, it does not project ForceP. Instead, the clause begins with TopP, which hosts the constituent *ʕali* in its Specifier position, Spec,TopP, as a topic. The derivation then proceeds through TP and the lower verbal domain vP, maintaining a full subject–predicate structure within a biclausal syntactic configuration.

**Negative:**

(7) qul-k lak mā jāʔ-š lil-bayt
say-1.SG to.you.MS NEG come.PST.3.MS-NEG to-the-house

‘I said to you that he didn’t come home’

The negation marker *lā...–š* in the Ibbi dialect functions as a bipartite negation strategy that applies across all tenses, in line with Shormani and Alhussen (2024). Example (7) demonstrates negation within *qulk*-embedded clauses, where *lā* precedes and –š follow the embedded verb *jāʔ*, preserving VSO word order. The post-verbal appearance of –š confirms its position in NegClP, structurally higher than vP but within the TP domain. The configuration is schematized in (8) below:

(8)

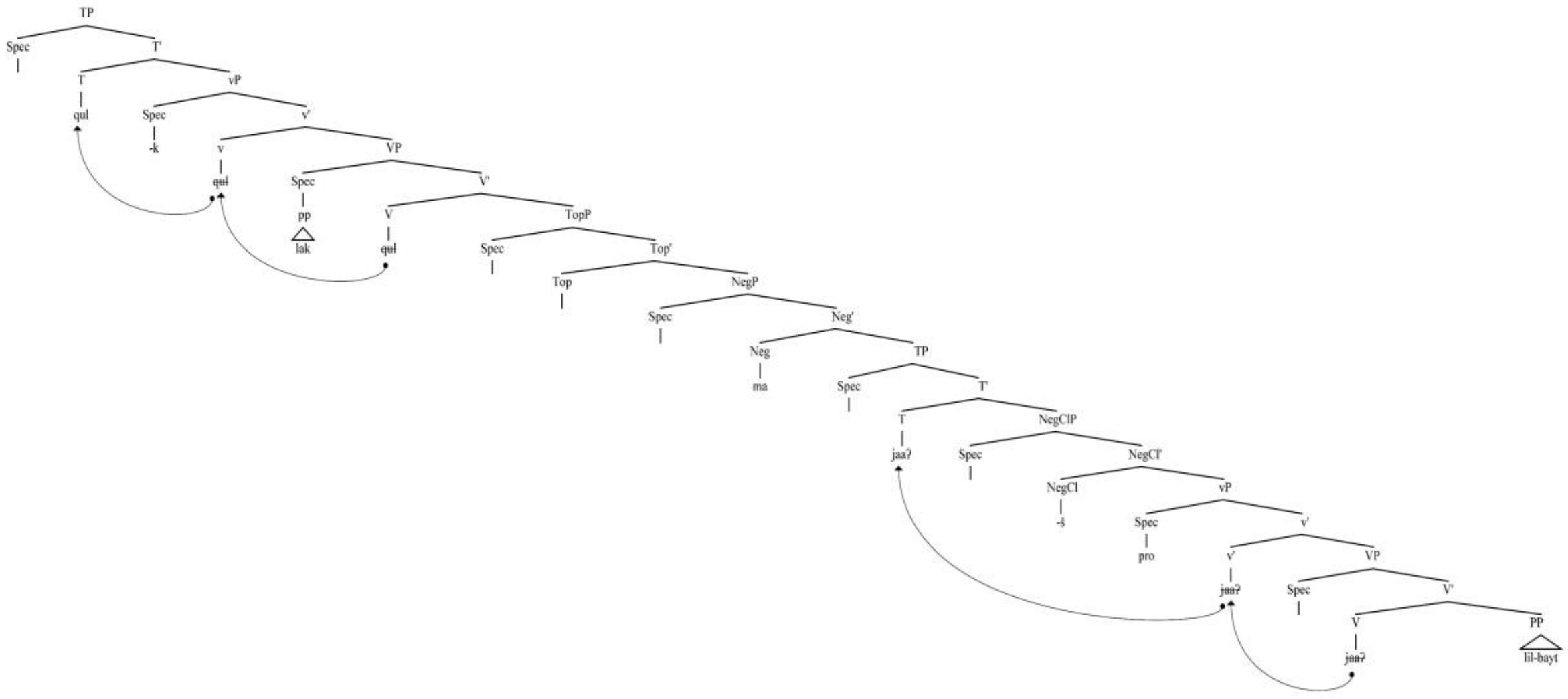

The sentence in (8) exemplifies the bipartite negation structure in YIA. Syntactically, the matrix clause is headed by *qulk* I said, where the verb *qul* moves to T, merging with the clitic *-k* via M-Merger (cf. Shormani, 2014; Matushansky, 2006). The embedded clause exhibits bipartite negation *mā...-š*, which projects as two distinct functional heads: NegP (hosting the preverbal ma) and NegClP, hosting the clitic -š, as proposed by Shormani & Alhussen (2024). The subject of the embedded clause is a null pro, licensed under Agree with T, while the verb jāʔ ‘came’ remains within TP. The prepositional phrase *lil-bayt* ‘to.the-house’ is base-generated in VP. This structure underscores YIA's adherence to minimalist derivational economy, with negation and topic interacting at the syntax-discourse interface. During the post-syntactic phase, the structure undergoes transfer to the phonological level (PF). Although the clitic *-k* and the verb *qul* are not syntactically adjacent, they enter a local head-adjacency configuration (Matushansky 2006, p. 86, 87) – specifically, a specifier-head relationship – which allows the morphological operation of M-Merger to apply (Matushansky 2006, p. 82, 95-96). This operation, which is part of an independent morphological component and not narrow syntax, fuses the two heads into a single, syntactically opaque node (Matushansky 2006, p. 95-96), forming a single phonological word, *qulk*, despite their original non-adjacency in the syntactic structure.

**Emphatic:**

(9) qul-k lak ʔinna ʕali jāʔ ʔams
say-1.SG to.you.MS that Ali come.PST.3.MS yesterday
'I said to you that Ali came yesterday.'

(10)

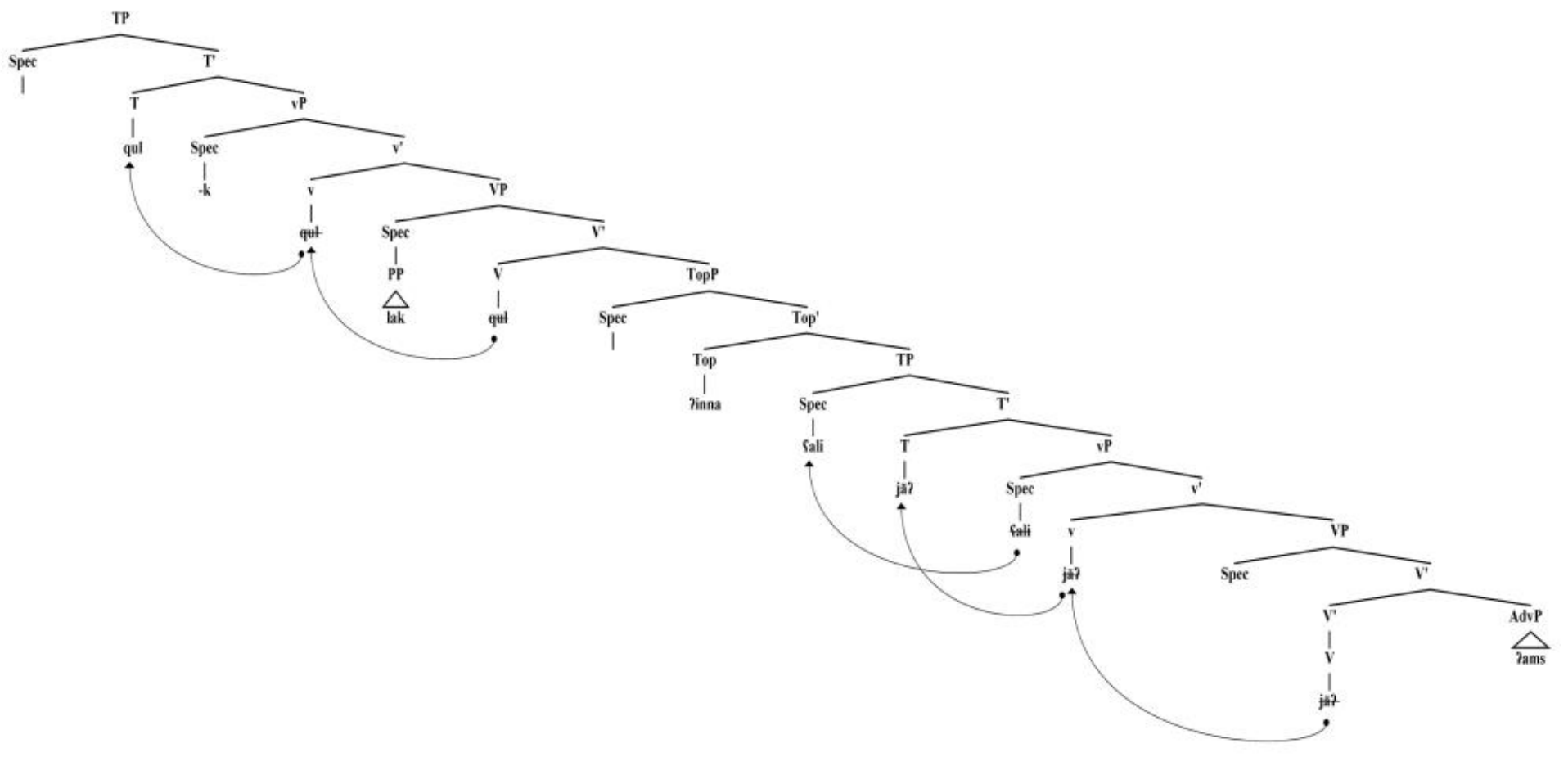

Details aside, the sentence in (10) combines a main clause (*qulk* lak) with an embedded declarative clause introduced by ʔinna. In this proposal, ʔanna occupies the head of TopP, marking the following clause *ʕali jāʔ ʔams* as a factual statement. The time adverb *ʔams* appears clause-finally, reflecting the dialect's typical VSO structure and the postverbal position of temporal modifiers.

**4.2. Interrogative *qulk*-clauses**

(11) a. qul-k wayn wali ʕali

say-1.SG where go.PST.3.MS Ali
‘I said: Where did Ali go?’

b. qul-k ʔayš ʔištara ʕali
say-1.SG what buy.PST.3.MS Ali
‘I said: What did Ali buy?’

The examples in (11a-b) show that *qul*-k introduces interrogative complements with wh-words *wayn* ‘where’, ʔayš ‘what’ in situ, maintaining VSO order and lacking an overt complementizer. The absence of structural differences between declarative and interrogative complements under *qul* indicates its uniform selectional properties across clause types. The syntactic analysis will be visually demonstrated in the tree diagram below, with detailed explanation of example in (12):

(12). qul-k wayn wali ʕali
say-1.SG where go.PST.3.MS Ali
‘I said: Where did Ali go?’

(13)

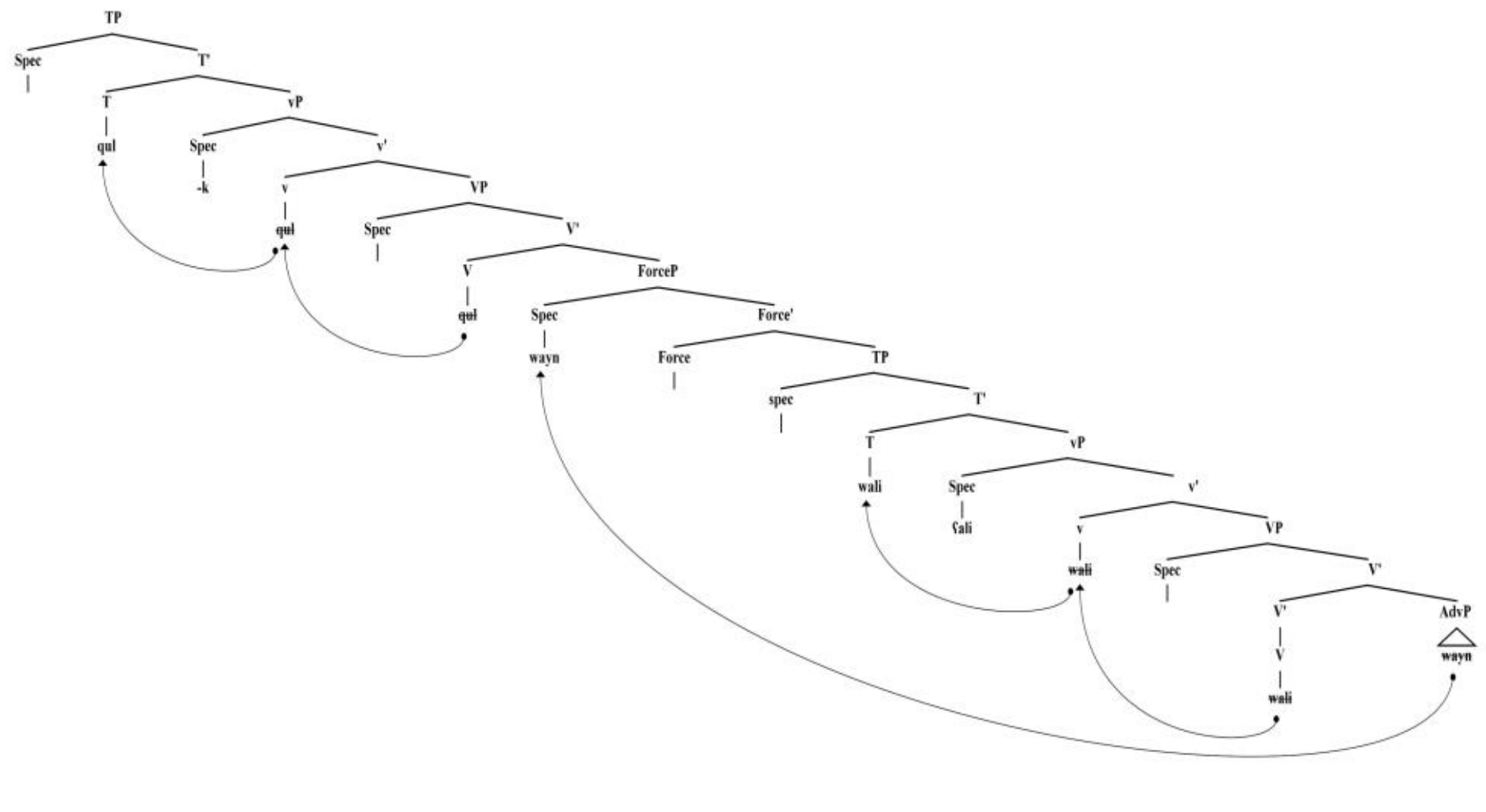

This sentence in (13) shows wh-movement in Ibbi Arabic, where the question operator *wayn* moves to the beginning of the embedded clause, specifically to Spec,ForceP, to form a question—just like in English. The verb *wali* stays in its base position after the subject *ʕali* because Arabic doesn’t require subject-verb inversion in questions. The complementizer position remains empty since no auxiliary movement is needed, keeping the structure minimal. The main clause *qulk* introduces the embedded question without affecting its internal word order, confirming that wh-words in Arabic must front but don't trigger additional verb movement (Shormani, 2024a)

### .4.3. Imperative *qulk*-clauses

Imperatives issue commands with or without negation. *qulk* softens directives:

**Affirmative Imperative**

(14) a. qul-k ʔiftaħ al-bāb
say-1.SG open.IMP.2.MS the-door
'I said: Open the door.'

b. qul-k lak hāt al-kitāb
say-1.SG to.you.MS bring.IMP.2.MS the-book
'I said to you: Bring the book,'

The imperative *qul-k* constructions in (14a-b) exhibit direct-speech encoding with second-person imperatives *ʔiftaħ*, *hāt* embedded under a matrix verb of saying. Crucially, the imperative verbs retain their canonical directive force despite being reported, reflecting their status as non-subordinate clauses with independent illocutionary strength. The construction underscores the imperative's resistance to syntactic embedding in typical complementation structures (Aikhenvald, 2010; Crnič & Trinh, 2009)), instead favoring direct quotation frames that preserve its deictic anchoring to the addressee (here, the implied *you* in (14a) and the overt *lak* 'to you' in (14b)).

**Negative Imperative**

(15) qul-k lak lā tiftaħ-š al-bāb
say-1.SG to.you.MS NEG open.IMP.2.MS-NEG the-door
'I said to you: Don't open the door.'

In YIA, negative imperatives follow a bipartite structure composed of the preverbal negative particle lā and the postverbal clitic -š, which together form the standard negation pattern for prohibitions. Rather than using only *lā* before the verb, the natural and grammatical form includes -š after the verb, resulting in *lā tiftaħ-š* instead of just *lā tiftaħ*. The element *lak* 'to you' serves a discourse function, identifying the addressee, while the actual syntactic subject is a covert pro, interpreted through discourse linkage (Shormani, 2024a).

The syntactic structure of the affirmative and negative imperative can be represented as in (17).

**Affirmative Imperative**

(16). qul-k lak hāt al-kitāb alʔān
say-1.SG to.you.MS bring.IMP.2.MS the-book now
'I said to you: Bring the book now,'

(17)

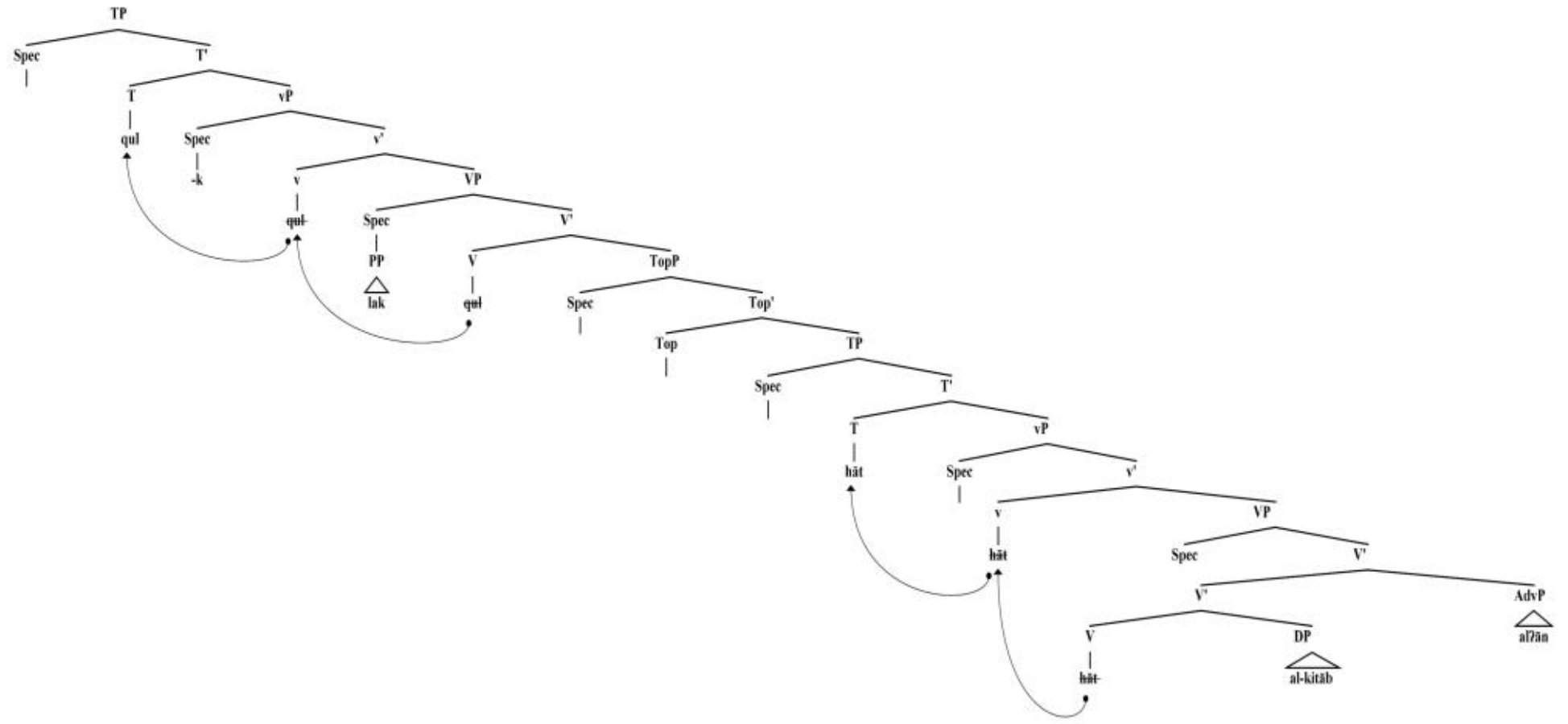

In the sentence (17), the verb hat originates in the V position and raises to a higher functional head within the embedded clause to satisfy its imperative interpretation. The subject an implied you remains covert. The matrix clause *qulk lak* introduces the embedded command without altering its internal structure.

**Negative Imperative**

(18) qul-k lak lā tiftaħ-š al-bāb
say-1.SG to.you.MS NEG open.IMP.2.MS-NEG the-door
'I said to you: Don't open the door.'

(19)

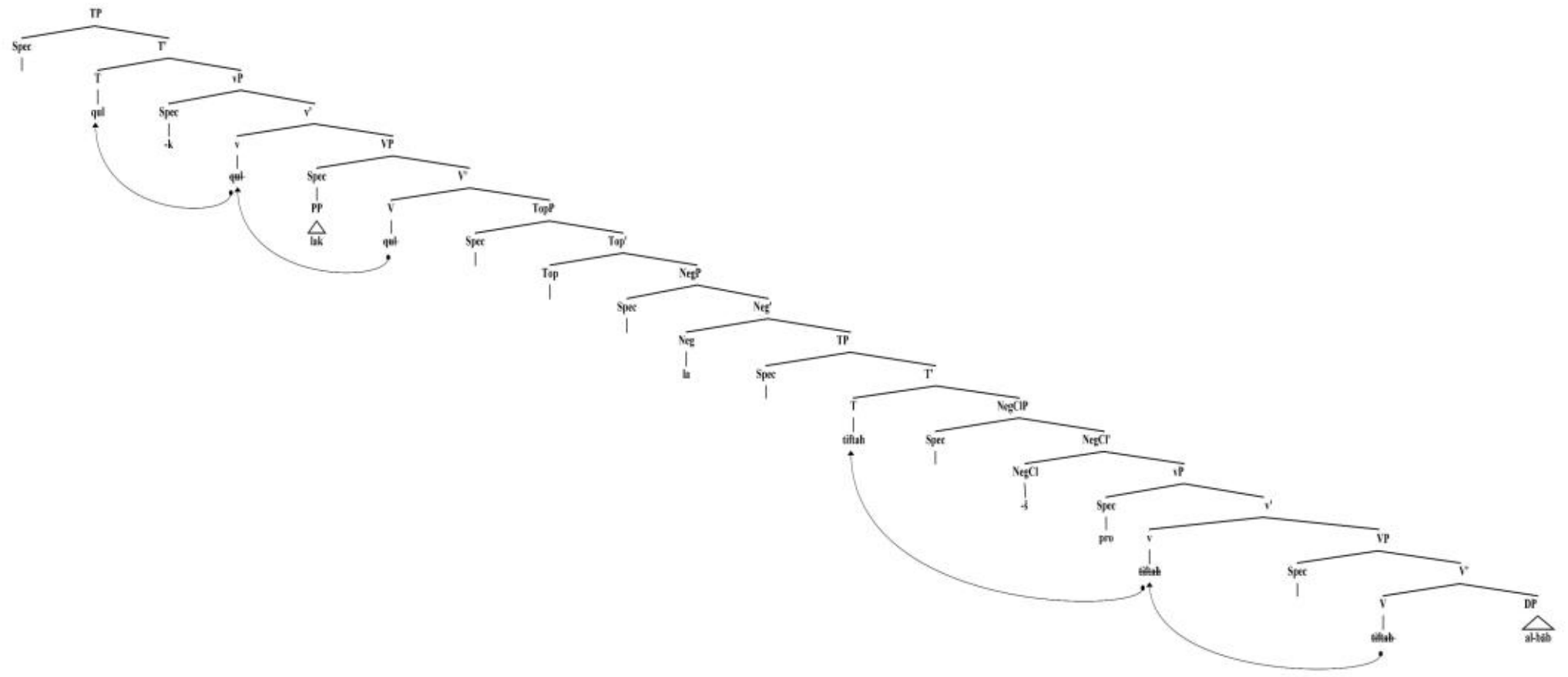

The structure in (19) begins with the matrix clause *qulk*. The embedded clause is a negative imperative, structured within a minimal left periphery. The verb *tiftaħ* raises from V through v⁰ to T⁰, with an implied subject in Spec,vP checked via Agree. Negation appears as bipartite, with lā in the head of NegP, and *-š* in NegClP (Shormani & Alhussen, 2024). The object *al-bāb* remains in situ within the VP. This structure preserves the VSO word order characteristic of Arabic while expressing negation through two distinct projections.

## 5. Conclusions and implications

This study has attempted to do more than just describe the structure of *qulk*-clauses in YIA; it shows how the grammar of an everyday spoken dialect reflects deep, systematic, and abstract principles of human language. What began as an exploration of a seemingly simple expression *qulk*-clause reveals a much richer reality: YIA speakers regularly and intuitively construct complex biclausal structures, complete with embedded propositions, clitic subject markers, and even nuanced patterns of negation and modality.

Thus, a number of conclusions can be drawn from our study: i) *qulk*-clauses in YIA can be formed with different and several types of clauses, we have employed only three, viz., declarative, question, and imperative, ii) these *qulk*-clauses are much similar to other structures including Topic-comment constructions (Shormani & Qarabesh, 2018), imperative structures (Shormani, 2021), and can have bipartite negations (Shormani & Alhussen, 2024), iii) there is a strong connection between syntax and discourse in both the derivation and processing, and interpretation of these structures, much like those of other constructions involving TP-domain and CP-domain, viz., the propositional domain and information domain, respectively. And iv) the PM adequately accounts for *qulk*-clauses in YIA. What is striking is, however, how closely the syntactic behavior of these clauses aligns with the PM assumptions and mechanisms, despite being informal and dialectical, unlike, say, Standard Arabic structures. This strongly suggests that these dialectical varieties of Arabic are no less rule-governed than Standard Arabic; they simply operate with similar UG principles and parameters. Thus, *qulk*-clauses in YIA provide further evidence in support of the PM (see Chomsky, 2021).

However, many questions remain unanswered. For instance, given that *qulk*-clauses are speaker-oriented, do addressee-clauses *qil-k* 'you said' behave the same way? Another very important issue concerns the acquisition of *qulk*-clauses. Given that *qulk*-clauses are complex in nature, i.e. they link CP-domain to TP-domain, how do children acquire them?, and more importantly, at what age? The answer to these, among other related, questions could offer valuable insights into the accessibility and cognitive salience of these structures, and perhaps the universality of minimalism. And we leave these for further studies.